\documentclass[anon,12pt]{clear2023} 

\usepackage{algpseudocode}
\usepackage{algorithm}

\algnewcommand\algorithmicforeach{\textbf{for each}}
\algdef{S}[FOR]{ForEach}[1]{\algorithmicforeach\ #1\ \algorithmicdo}

\usepackage[most]{tcolorbox}
\usepackage{multicol}
\usepackage{hhline}
\usepackage{tensor}
\usepackage{soul}
\newcommand{\hlc}[2][yellow]{{%
    \colorlet{foo}{#1}%
    \sethlcolor{foo}\hl{#2}}%
}

\usepackage{amsthm}
\makeatletter
\newtheorem*{rep@theorem}{\rep@title}
\newcommand{\newreptheorem}[2]{%
\newenvironment{rep#1}[1]{%
 \def\rep@title{#2 \ref{##1}}%
 \begin{rep@theorem}}%
 {\end{rep@theorem}}}
\makeatother

\newtheorem{assumptions1}{Assumption}
\newtheorem{theorem1}{Theorem}
\newreptheorem{theorem1}{Theorem}
\newtheorem{lemma1}{Lemma}
\newreptheorem{lemma1}{Lemma}
\newtheorem{corollary1}{Corollary}
\newreptheorem{corollary1}{Corollary}
\newtheorem{proposition1}{Proposition}
\newreptheorem{proposition1}{Proposition}
\newtheorem{definition1}{Definition}

\usepackage[labelformat=simple]{subcaption}

\DeclareMathOperator*{\argmin}{arg\,min}

\usepackage{colortbl}

\usepackage{tikz}
\usepackage{pgfplots}
\usepackage{rotating} 
\usepackage{enumitem}
\usetikzlibrary{decorations.pathreplacing,arrows,automata,calc,shapes.geometric}

\pgfdeclaredecoration{lightning bolt}{draw}{
\state{draw}[width=\pgfdecoratedpathlength]{
  \pgfpathmoveto{\pgfpointorigin}%
  \pgfpathlineto{\pgfpoint{\pgfdecoratedpathlength*0.6}%
    {-\pgfdecoratedpathlength*.1}}%
  \pgfpathlineto{\pgfpoint{\pgfdecoratedpathlength*0.55}{0pt}}%
  \pgfpathlineto{\pgfpoint{\pgfdecoratedpathlength}{0pt}}%
  \pgfpathlineto{\pgfpoint{\pgfdecoratedpathlength*0.4}%
    {\pgfdecoratedpathlength*.1}}%
  \pgfpathlineto{\pgfpoint{\pgfdecoratedpathlength*0.45}{0pt}}%
  \pgfpathclose%
}%
}

\usepackage{bm}
\newcommand\ci{\perp\!\!\!\perp}

\makeatother

\makeatletter
\def\SOUL@hlpreamble{%
    \setul{\dp\strutbox}{\dimexpr\ht\strutbox+\dp\strutbox\relax}%
    \let\SOUL@stcolor\SOUL@hlcolor
    \SOUL@stpreamble
}
\makeatother

\let\oldnl\nl
\newcommand{\nonl}{\renewcommand{\nl}{\let\nl\oldnl}}

\title[Extract Errors with Latents]{Sample-Specific Root Causal Inference with Latent Variables}
\usepackage{times}

\clearauthor{%
 \Name{Author Name1} \Email{abc@sample.com}\\
 \addr Address 1
 \AND
 \Name{Author Name2} \Email{xyz@sample.com}\\
 \addr Address 2%
}

\begin{document}

\maketitle

\begin{abstract}%
  Root causal analysis seeks to identify the set of initial perturbations that induce an unwanted outcome. In prior work, we defined sample-specific root causes of disease using exogenous error terms that predict a diagnosis in a structural equation model. We rigorously quantified predictivity using Shapley values. However, the associated algorithms for inferring root causes assume no latent confounding. We relax this assumption by permitting confounding among the predictors. We then introduce a corresponding procedure called Extract Errors with Latents (EEL) for recovering the error terms up to contamination by vertices on certain paths under the linear non-Gaussian acyclic model. EEL also identifies the smallest sets of dependent errors for fast computation of the Shapley values. The algorithm bypasses the hard problem of estimating the underlying causal graph in both cases. Experiments highlight the superior accuracy and robustness of EEL relative to its predecessors.
\end{abstract}


\section{Introduction}

Causal inference refers to the process of inferring causal relations from data. Most scientists identify causal relations by conducting randomized controlled trials (RCTs). RCTs can nevertheless fail to distinguish between a cause and a \textit{root cause} of disease, or the initial perturbation to an otherwise healthy system that ultimately induces a diagnostic label. Identifying root causes is critical for (a) understanding disease mechanisms and (b) discovering drug targets that eliminate disease \textit{at its onset} in a biological pathway.

Consider for example the directed graph in Figure \ref{fig_shock}, where vertices in $\bm{X}$ represent random variables and directed edges their direct causal relations; we have $X_i \rightarrow X_j$ when $X_i$ directly causes $X_j$. The lightning bolt in the figure denotes an exogenous perturbation of the root cause $X_2$, such as a virus, mutation or physical injury. This perturbation in turn affects many downstream variables, such as $\{X_3, X_4\}$, ultimately causing symptoms $\{X_5, X_6\}$ and physicians to label a patient with a diagnosis $D=1$ indicating disease. The causes of $D$ include $X_1, \dots, X_6$, but we only seek to identify the root cause $X_2$ that may lie arbitrarily far upstream from $D$ in the general case.

Identifying root causes is further complicated by the existence of \textit{complex disease}, where each patient may have multiple root causes, and root causes may differ between patients even within the same diagnostic category. The disease may also only affect certain tissues or cells in the body. We therefore more specifically seek to identify \textit{sample-specific} root causes, where a sample may denote an arbitrary unit of granularity such as a patient, tissue or cell. Identifying sample-specific root causes has the potential to help experimentalists rapidly identify interventions that target the very beginnings of disease unique to each patient. 

The above intuitive idea of a sample-specific root cause nevertheless lacks a rigorous mathematical definition. This in turn hinders the development of principled algorithms designed for their automated detection. As a result, we explicitly defined sample-specific root causes of disease as the error terms in a structural equation model that predict a diagnostic label in prior work \citep{Strobl22a}. We quantified predictivity using Shapley values. We also proposed methods to directly extract these error terms both in the linear and non-linear settings via regression residuals \citep{Strobl22a,Strobl22b}. The methods do not require knowledge about the underlying graph and achieve sample efficiency by bypassing the hard problem of causal graph recovery \citep{Chickering04}. These algorithms however rely on the unreasonable assumption that the dataset contain no unobserved confounders, which we relax in this paper by permitting confounding between the variables in $\bm{X}$.

\begin{figure}
\centering
\begin{subfigure}{0.49\textwidth}
\centering
\raisebox{1.0cm}{\begin{tikzpicture}[scale=1.0, shorten >=1pt,auto,node distance=2.8cm, semithick,
  inj/.pic = {\draw (0,0) -- ++ (0,2mm) 
                node[minimum size=2mm, fill=red!60,above] {}
                node[draw, semithick, minimum width=2mm, minimum height=5mm,above] (aux) {};
              \draw[thick] (aux.west) -- (aux.east); 
              \draw[thick,{Bar[width=2mm]}-{Hooks[width=4mm]}] (aux.center) -- ++ (0,4mm) coordinate (-inj);
              }]
                    
\tikzset{vertex/.style = {inner sep=0.4pt}}
\tikzset{edge/.style = {->,> = latex'}}
 
\node[vertex] (1) at  (0,0) {$X_1$};
\node[vertex] (2) at  (1.25,0) {$X_2$};
\node[vertex] (3) at  (2.5,0.5) {$X_3$};
\node[vertex] (4) at  (2.5,-0.5) {$X_4$};
\node[vertex] (5) at  (3.75,0.5) {$X_5$};
\node[vertex] (6) at  (3.75,-0.5) {$X_6$};
\node[vertex] (7) at  (5,0) {$D$};

\fill [blue, decoration=lightning bolt, decorate] (1.25,0.25) -- ++ (0.55,0.65);

\draw[edge] (1) to (2);
\draw[edge,blue] (2) to (3);
\draw[edge,blue] (2) to (4);
\draw[edge,blue] (3) to (5);
\draw[edge,blue] (4) to (6);
\draw[edge,blue] (5) to (7);
\draw[edge,blue] (6) to (7);
\end{tikzpicture}}
\caption{} \label{fig_shock}
\end{subfigure} \begin{subfigure}{0.49\textwidth}
\centering
\begin{tikzpicture}[scale=1.0, shorten >=1pt,auto,node distance=2.8cm, semithick]
                    
\tikzset{vertex/.style = {inner sep=0.4pt}}
\tikzset{edge/.style = {->,> = latex'}}
 
\node[vertex] (1) at  (0,0) {$X_1$};
\node[vertex,draw=blue, circle] (2) at  (1.25,0) {$X_2$};
\node[vertex,draw=blue, circle] (3) at  (2.5,0.5) {$X_3$};
\node[vertex,draw=blue, circle] (4) at  (2.5,-0.5) {$X_4$};
\node[vertex,draw=blue, circle] (5) at  (3.75,0.5) {$X_5$};
\node[vertex,draw=blue, circle] (6) at  (3.75,-0.5) {$X_6$};
\node[vertex,draw=blue, circle, minimum size=0.55cm] (7) at  (5,0) {$D$};

\node[vertex] (8) at  (-0.5,1) {$E_1$};
\draw[edge] (8) to (1);
\node[vertex] (9) at  (0.75,1) {$E_2 = \textcolor{blue}{e_2}$};
\draw[edge,blue] (9) to (2);
\node[vertex] (10) at  (2,1.5) {$E_3$};
\draw[edge] (10) to (3);
\node[vertex] (11) at  (3.25,1.5) {$E_5$};
\draw[edge] (11) to (5);
\node[vertex] (13) at  (2,-1.5) {$E_4$};
\draw[edge] (13) to (4);
\node[vertex] (14) at  (3.25,-1.5) {$E_6$};
\draw[edge] (14) to (6);

\draw[edge] (1) to (2);
\draw[edge,blue] (2) to (3);
\draw[edge,blue] (2) to (4);
\draw[edge,blue] (3) to (5);
\draw[edge,blue] (4) to (6);
\draw[edge,blue] (5) to (7);
\draw[edge,blue] (6) to (7);
\end{tikzpicture}
\caption{} \label{fig_SEM_down}
\end{subfigure}
\caption{The lightning bolt in (a) denotes an exogenous perturbation of $X_2$ that eventually affects many downstream variables and causes a diagnosis $D$. In (b), we model the lightning bolt as a perturbation of $E_2$ to the value $e_2$ that impacts the values of all of its descendants and ultimately $D$.}
\end{figure}
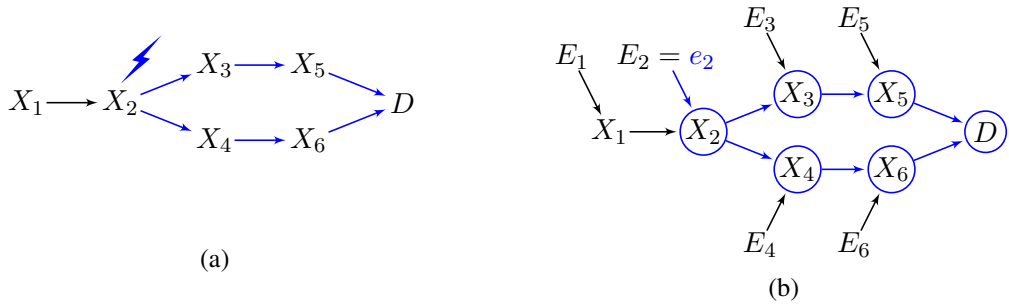
\begin{tcolorbox}[breakable,enhanced,frame hidden]
We specifically make the following contributions in this paper:
\begin{itemize}[leftmargin=*]
\item We introduce a strategy for identifying sample-specific root causes with confounding by extracting the error terms up to contamination on certain paths.
\item We propose an algorithm called Extract Errors with Latents (EEL) that recovers the above error terms and computes an undirected graph summarizing their statistical dependencies.
\item We use the graph to efficiently compute Shapley values of the error terms by averaging over small neighborhoods of dependence. 
\end{itemize}
\end{tcolorbox}
\noindent Experiments in Section \ref{sec:exp} highlight the superiority of EEL relative to existing methods in the presence of confounding.

\section{Structural Equation Models}
We can formalize causal inference under the framework of structural equation models (SEMs). An SEM over a set of $p$ random variables $\bm{X}$ refers to a set of deterministic equations in the following form:
\begin{equation} \nonumber
    X_i = f_i(\textnormal{Pa}(X_i), E_i), \hspace{3mm} \forall X_i \in \bm{X}.
\end{equation}
where $\bm{E}$ denotes a random vector of $p$ mutually independent error terms, and $\textnormal{Pa}(X_i) \subseteq \bm{X}$ the \textit{parents}, or direct causes, of $X_i$. We can equivalently set the equality sign in the above equation to algorithmic assignment $\leftarrow$ in order to emphasize that interventions on $\textnormal{Pa}(X_i)$ induce changes in the conditional probability distribution of $X_i$ given $\textnormal{Pa}(X_i)$. 

We can associate an SEM with a \textit{directed graph} $\mathbb{G}$ containing at most one directed edge between any two variables in $\bm{X}$. We have $X_i \rightarrow X_j$, when there exists a direct causal relation from $X_i \in \textnormal{Pa}(X_j)$ to $X_j$. We always have $E_i \rightarrow X_i$ in $\mathbb{G}$ but only draw the vertices in $\bm{E}$ and their outgoing edges when informative. We use the notation $\textnormal{Pa}_{\mathbb{G}}(X_j)$ to emphasize the underlying graph $\mathbb{G}$. A \textit{directed path} is sequence of directed edges. $X_i$ is an \textit{ancestor} of $X_j$, and $X_j$ a \textit{descendant} of $X_i$, if a directed path exists from $X_i$ to $X_j$. A \textit{directed acyclic graph} (DAG) corresponds to a directed graph without \textit{cycles}, where $X_i$ is an ancestor of $X_j$ and vice versa. A vertex $X_j$ is a \textit{collider} on a path if we have $X_i \rightarrow X_j \leftarrow X_k$ on the path. Two vertices $X_i$ and $X_j$ are \textit{d-connected} given $\bm{W} \setminus \{X_i,X_j\}$ when there exists a path between $X_i$ and $X_j$ such that every collider has a descendant in $\bm{W}$ and no non-collider is in $\bm{W}$. The two vertice are likewise \textit{d-separated} when they are not d-connected.

An SEM with an associated DAG $\mathbb{G}$ can admit a density that factorizes according to the graph:
\begin{equation} \nonumber
p(\bm{X}) = \prod_{i=1}^p p(X_i | \textnormal{Pa}_{\mathbb{G}}(X_i)).
\end{equation}
The above factorization implies that, if $X_i$ and $X_j$ are d-separated given $\bm{W}$ in $\mathbb{G}$, then the two vertices are also conditionally independent given $\bm{W}$, which we denote by $X_i \ci X_j | \bm{W}$ for shorthand \citep{Lauritzen90}. \textit{D-separation faithfulness} refers to the converse: if $X_i \ci X_j | \bm{W}$, then $X_i$ and $X_j$ are d-separated given $\bm{W}$.

In this paper, we focus on linear SEMs with an associated DAG:
\begin{equation} \label{eq:SEM_lin}
    X_i = \sum_{j=1}^p X_j \beta_{ji} + E_i, \hspace{3mm} \forall X_i \in \bm{X},
\end{equation}
comprised of a set of linear equations with coefficient matrix $\beta$ where $\beta_{ji} \not = 0$ if and only if $X_j \in \textnormal{Pa}_{\mathbb{G}}(X_i)$. We assume $\mathbb{E}(\bm{X}) = 0$ without loss of generality. The equations more specifically follow a \textit{Linear Non-Gaussian Acyclic Model} (LiNGAM) when each error term is continuous non-Gaussian \citep{Shimizu06}.

Most existing methods also assume that we observe all of the variables in $\bm{X}$. We relax this assumption by dividing $\bm{X}$ into a set of $q$ observed variables $\bm{O}$ and a set of $m$ latent -- or unobserved -- common causes $\bm{L}$. We can \textit{always} write the following:
\begin{equation} \label{eq:SEM_lat}
    O_i = \sum_{j=1}^q O_j \beta_{ji} + \sum_{k=1}^m L_k \gamma_{ki} + E_i, \hspace{3mm} \forall O_i \in \bm{O}.
\end{equation}
Each $L_k$ must have at least two children lest we accommodate it into $E_i$. Without loss of generality, we may also assume that $\bm{T} = \bm{L} \cup \bm{E}$ denotes a set of mutually independent random variables with no parents \citep{Hoyer08}. We refer to Equation \eqref{eq:SEM_lat} as the \textit{canonical form}.

We can write Equation \eqref{eq:SEM_lat} in matrix notation:
\begin{equation} \nonumber
    \bm{O} = \bm{O} \beta + \bm{L} \gamma + \bm{E}.
\end{equation}
Re-arranging terms yields:
\begin{equation} \nonumber
    \bm{O} = (\bm{L} \gamma + \bm{E})(I - \beta )^{-1}
    = \bm{E} \lambda + \bm{L} \gamma\lambda = \bm{T} \theta\\
\end{equation}
where $\lambda = (I - \beta )^{-1}$ and $\theta = [\lambda; \gamma\lambda]$. Notice that $\bm{T}$ is now ordered such that $T_j = E_j$ if $j \leq q$. The entry $\theta_{ji}$ quantifies the \textit{total effect} of the latent variable or error term $T_j$ on $O_i$. 

\section{Sample-Specific Root Causes}

We consider LiNGAM over $\bm{X}$ and introduce an additional label $D$ representing a diagnosis; we have $D=1$ for patients deemed to have a disease, and $D=0$ for healthy controls. We then assume a DAG over $\bm{X} \cup D$ such that $D$ is a \textit{terminal vertex}, or a vertex without descendants, and linked to $\bm{X}$ via a logistic function:
\begin{assumptions1} \label{assump_terminal}
$D$ is a terminal vertex such that $\mathbb{P}(D|\bm{X}) =\ \textnormal{logistic}(\bm{X}\beta_{\cdot D} + \alpha)$.
\end{assumptions1}
\noindent This is a reasonable assumption because a scientist who seeks to identify the causes of $D$ will likely use datasets containing measurements of the non-descendants of the diagnosis, such as gene expression levels, clinical laboratory values or imaging. The logistic link also provides a natural extension of LiNGAM to handle a noisy binary variable.

We model a sample-specific perturbation first affecting the root cause $X_i \in \bm{X}$ as a change in the value of its error term $E_i$. We may write the following for any healthy control:
\begin{equation}
X_i = \sum_{j=1}^p X_j \beta_{ji} + \widetilde{e}_i,
\end{equation}
where we have set the value of $E_i$ in Equation \eqref{eq:SEM_lin} to $\widetilde{e}_i$. Suppose however that an exogenous perturbation -- such as a virus, mutation or physical injury -- changes the value of $E_i$ from $\widetilde{e}_i$ to $e_i$. This perturbation in turn effects the value of $X_i$ and all of its downstream effects, ultimately increasing the probability of developing disease $D=1$ (Figure \ref{fig_SEM_down}).

We can quantify the increase in the probability of developing disease using logistic regression. We in particular consider:
\begin{equation} \nonumber
f(\bm{E}) = \textnormal{ln} \Big[\frac{\mathbb{P}(D=1| \bm{E})}{\mathbb{P}(D=0|\bm{E})}\Big] = \bm{E} \theta_{\cdot D} + \alpha,
\end{equation}
where the last equality follows by Assumption \ref{assump_terminal}. Let $v(\bm{W})$ denote the conditional expectation of the logistic regression model $\mathbb{E}(f(\bm{E})|\bm{W})$, initially where $\bm{W} = \emptyset$. We then measure the change in probability when intervening on $E_i$ via the following difference:
\begin{equation} \label{eq_emptyW}
\gamma_{E_i \bm{W}} = \underbrace{v(E_i,\bm{W})}_{(a)}-\underbrace{v(\bm{W})}_{(b)}
\end{equation}
We have $\gamma_{e_i \bm{W}} >0$ when $E_i = e_i$ increases the probability that $D=1$ because (a) is larger than (b).

Expression \eqref{eq_emptyW} unfortunately only quantifies the effect of $E_i$ on $D$ \textit{in isolation}. We however also want to quantify the joint effect of $E_i$ in conjunction with the other error terms in $\bm{E} \setminus E_i$ when $\bm{W} \not = \emptyset$. We therefore average over all possible combinations of the errors as follows:
\begin{equation} \label{eq_shapley}
    S_i = \frac{1}{p}\hspace{-15mm}\underbrace{\sum_{\bm{W} \subseteq (\bm{E} \setminus E_i)} \frac{1}{\binom{p-1} {|\bm{W}|}}}_{\textnormal{Average over all possible combinations of } \bm{E} \setminus E_i} \hspace{-14.5mm}\gamma_{E_i \bm{W}}.
\end{equation}
The quantity corresponds precisely to the well-known Shapley value which, as the reader may recall, is the \textit{only} value satisfying the linearity, efficiency, symmetry and null player properties (see e.g., \citep{Lundberg17,Strumbelj15}).

The following result holds:
\begin{proposition1} \label{prop:total}
Under LINGAM over $\bm{X}$ and Assumption \ref{assump_terminal}, the Shapley value $S_i$ corresponds to the sample-specific total effect of $E_i$ on $D$: $S_i = E_i \theta_{iD}$.\footnote{If $D = \bm{X} \beta_{\cdot D} + E_D$ is terminal and continuous, then we arrive at the same result when $\gamma_{E_i \bm{W}} = \mathbb{E}(D|E_i,\bm{W}) - \mathbb{E}(D|\bm{W})$. We focus on a binary target because this is the most common situation encountered by far.}
\end{proposition1}
\noindent The proof follows directly from Corollary 1 of \citep{Lundberg17}. This justifies the following definition of a sample-specific root cause:
\begin{definition1} \label{def:root}
$X_i \in \textnormal{Anc}_{\mathbb{G}}(D)$ is a \textit{sample-specific root cause of disease} if $S_i > 0$.
\end{definition1}
\noindent In other words, a sample-specific root cause of disease is a variable associated with an error term that increases the probability that $D=1$ as quantified by the Shapley value $S_i > 0$. We do not consider $S_i \leq 0$ because $E_i$ decreases the probability that $D=1$ (or likewise increases the probability that $D=0$) when $S_i < 0$. Similarly, $E_i$ has no effect on increasing or decreasing the probability that $D=1$ when $S_i = 0$. We have thus arrived at a concise definition of a sample-specific root cause as a variable associated with a positive Shapley value of its error.

\section{Inducing Paths \& Terms}

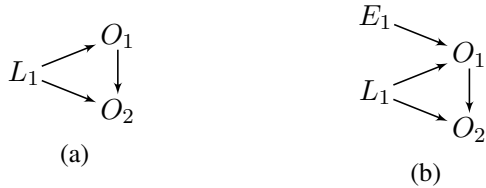
\begin{figure}[t]
\centering
\begin{subfigure}{0.3\textwidth}
\centering
\begin{tikzpicture}[scale=1.0, shorten >=1pt,auto,node distance=2.8cm, semithick,
  inj/.pic = {\draw (0,0) -- ++ (0,2mm) 
                node[minimum size=2mm, fill=red!60,above] {}
                node[draw, semithick, minimum width=2mm, minimum height=5mm,above] (aux) {};
              \draw[thick] (aux.west) -- (aux.east); 
              \draw[thick,{Bar[width=2mm]}-{Hooks[width=4mm]}] (aux.center) -- ++ (0,4mm) coordinate (-inj);
              }]
                    
\tikzset{vertex/.style = {inner sep=0.4pt}}
\tikzset{edge/.style = {->,> = latex'}}
 
\node[vertex] (2) at  (1.25,0) {$L_1$};
\node[vertex] (3) at  (2.5,0.5) {$O_1$};
\node[vertex] (4) at  (2.5,-0.5) {$O_2$};

\draw[edge] (2) to (3);
\draw[edge] (2) to (4);
\draw[edge] (3) to (4);
\end{tikzpicture}
\caption{} \label{fig_error_inexact:example}
\end{subfigure} \begin{subfigure}{0.3\textwidth} \centering
\begin{tikzpicture}[scale=1.0, shorten >=1pt,auto,node distance=2.8cm, semithick,
  inj/.pic = {\draw (0,0) -- ++ (0,2mm) 
                node[minimum size=2mm, fill=red!60,above] {}
                node[draw, semithick, minimum width=2mm, minimum height=5mm,above] (aux) {};
              \draw[thick] (aux.west) -- (aux.east); 
              \draw[thick,{Bar[width=2mm]}-{Hooks[width=4mm]}] (aux.center) -- ++ (0,4mm) coordinate (-inj);
              }]
                    
\tikzset{vertex/.style = {inner sep=0.4pt}}
\tikzset{edge/.style = {->,> = latex'}}

\node[vertex] (1) at (1.25,1) {$E_1$};
\node[vertex] (2) at  (1.25,0) {$L_1$};
\node[vertex] (3) at  (2.5,0.5) {$O_1$};
\node[vertex] (4) at  (2.5,-0.5) {$O_2$};

\draw[edge] (1) to (3);
\draw[edge] (2) to (3);
\draw[edge] (2) to (4);
\draw[edge] (3) to (4);
\end{tikzpicture}
\caption{} \label{fig_error_inexact:exampleIP}
\end{subfigure}

\caption{(a) Example where we cannot recover $E_1$ and $E_2$ exactly. (b) $E_1, L_1$ and $E_2$ each lie on a directed inducing path to $O_2$.} \label{fig_error_inexact}
\end{figure}

The definition of a sample-specific root cause implies that we must develop methods that can accurately extract the error terms in order to compute the Shapley value. We however cannot identify the error terms $\bm{E}$ exactly when confounding exists. Consider for example the graph shown in Figure \ref{fig_error_inexact:example}, where we cannot partial out $L_1$ from $O_1$ and $O_2$ because $L_1$ is unobserved. 

We can however identify the error terms up to connection by directed inducing paths:
\begin{definition1}
A \textit{directed inducing path} to $O_i$ is a path between $O_i$ and $T_j \in \bm{T}$ (possibly $i=j$) where every collider is an ancestor of $O_i$ and every non-collider is in $\bm{L}$.
\end{definition1}
\noindent All colliders are directed to $O_i$. We only consider directed inducing paths from the \textit{error terms} or \textit{latent variables} to $O_i$. We provide an example in Figure \ref{fig_error_inexact:exampleIP}. Any error term incident on or latent variable lying on a directed inducing path to $O_i$ also has a directed inducing path to $O_i$. Only $E_i$ lies on a directed inducing path to $O_i$ in the unconfounded setting, but more error terms may lie on the path when confounding exists. Finally, the above definition corresponds to the directed analogue of an (undirected) \textit{inducing path} utilized in constraint-based search with latent variables, where every collider is an ancestor of either endpoint $O_i$ or $T_j$ (or both) \citep{Spirtes00}.

The following result elucidates the limits of error term recovery when assessing statistical independence with regression residuals. Consider the ideal scenario where we have access to $F_j = E_j + \sum_{L_k \in \textnormal{Pa}_{\mathbb{G}}(O_j)\cap\bm{L}} L_k \gamma_{kj}$ for each $O_j \in \bm{O}$, which we collect into the set $\bm{F}$.
\begin{lemma1} \label{lem:IPs}
Under LiNGAM and d-separation faithfulness, if some entry in $\bm{W} \subseteq \bm{F} \setminus F_i$ corresponds to an observed vertex lying on a directed inducing path to $O_i$, then $R_{O_i \bm{W}} \not \ci F_j$ for some $F_j \in \bm{W}$.
\end{lemma1}
\noindent We delegate proofs to Appendix \ref{app:proofs}. The latent common causes lying on a directed inducing path to $O_i$ thus ensure that we cannot partial out the error terms incident on the path from $O_i$ in general, even if we identified all entries in $\bm{F} \setminus F_i$.

We instead focus on identifying the error terms \textit{up to} connection by a directed inducing path. Specifically, let $\bm{C}_i \subseteq \bm{T}$ denote the set of error terms and latent variables lying on any directed inducing path to $O_i$. We consider: 
\begin{equation} \label{eq:C_total}
E_i^* = \bm{C}_i \theta_{\bm{C}_ii},
\end{equation}
\noindent for each $O_i \in \bm{O}$. For example, $E_1^* = L_1\gamma_{11} + E_1$ and $E_2^* = E_1 \beta_{12} + L_1(\gamma_{12} + \gamma_{11}\beta_{12}) + E_2$ in Figure \ref{fig_error_inexact}. This generalizes the unconfounded setting where $E_i^* = E_i \theta_{\bm{C}_i i} = E_i$ because we have $\bm{C}_i = E_i$ and $\theta_{E_i i} = 1$ in this case. We call the set $\bm{E}^*$ the \textit{inducing terms}. The variable $E_i^*$ represents a corrupted estimate of the original error term $E_i$ in the sense that $E_i^*$ is a linear combination of $E_i$ and a small set of error terms and latent variables ancestral to $O_i$.

\section{Extracting Inducing Terms} \label{sec:EEL}

We now design an algorithm that identifies the inducing terms from the joint distribution of $\bm{O}$, without access to the ground truth DAG. We specifically build upon the DirectLiNGAM and EE algorithms explicated in Appendices \ref{app:DL} and \ref{app:EE} to handle cases where $\bm{L} \not = \emptyset$. 

We identify the inducing term of $O_j$ by regressing out as many of its ancestors in $\bm{T} \setminus E_j$ as possible. Let $\bm{W}$ denote a set of arbitrary linear combinations of error terms and latent variables in a minimal set $\bm{S} \subseteq \bm{T}$. The notation $R_{O_j\bm{W}}$ denotes the residuals of $O_j$ when linearly regressed on $\bm{W}$. We have:
\begin{proposition1} \label{prop:C}
    Under LiNGAM, $W_i$ is independent of the residuals $R_{O_j \bm{W}}$ for all $W_i \in \bm{W}$ if and only if $O_j$ can be written as a linear function of $\bm{W}$ \ul{plus} a linear function of $\bm{T} \setminus \bm{S}$. Thus, the residuals are a linear function of $\bm{T} \setminus \bm{S}$.
\end{proposition1}
\noindent The above proposition suggests that we should design an algorithm that iteratively replaces $O_j$ with $R_{O_j \bm{W}}$ because $R_{O_j \bm{W}}$ depends on a fewer number of members in $\bm{T}$. We can also partial out ancestors by performing a series of \textit{univariate and multivariate} regressions, progressively increasing the conditioning set size of $\bm{W}$. We partial out $\bm{W}$ from $O_j$ once we find a large enough $\bm{W}$ such that $R_{O_j \bm{W}} \ci O_i$ for all $O_i \in \bm{W}$.

Extract Errors with Latents (EEL) summarized in Algorithm \ref{alg_EEL} repeats the above procedure for each $O_j \in \bm{O}$. EEL proceeds just like EE but with additional steps highlighted in gray for increasing the conditioning set size. The algorithm first instantiates a complete undirected graph $\mathcal{G}$ over $\bm{O}$ in Line \ref{alg_EEL:graph}. The graph represents the statistical dependencies between the vertices in each iteration of the algorithm; $\mathcal{G}$ is not the DAG $\mathbb{G}$. The notation $\textnormal{Adj}_\mathcal{G}(O_j)$ refers the observed variables adjacent to $O_j$ in $\mathcal{G}$. The variable $l$ denotes the size of the set $\bm{W}$. EEL gradually increases $l$ until it finds a set $\bm{W} \subseteq \textnormal{Adj}_\mathcal{G}(O_j)$ where  $R_{O_j\bm{W}} \ci O_i$ for all $O_i \in \bm{W}$ in Line \ref{alg_EEL:indep}. EEL then partials out $\bm{W}$ from $O_j$ and removes the corresponding adjacencies from $\mathcal{G}$ in Lines \ref{alg_EEL:partial}-\ref{alg_EEL:adj}. The algorithm finally resets the size of $\bm{W}$ in Line \ref{alg_EEL:reset} by assigning $l \leftarrow 0$. This ensures that EEL proceeds with a fresh search after partialing out $\bm{W}$ from $O_j$.  

EEL recovers the inducing terms in the oracle setting. We first require a new definition:
\begin{definition1}
A \textit{confounding path} of $O_i$ is a path between $T_j \in \bm{T}$ and $O_k$ (possibly $i = j = k$) where every collider is an ancestor of $O_i$ and every non-collider is in $\bm{L}$.
\end{definition1}
\noindent A directed inducing path to $O_i$ must end at $O_i$, whereas a confounding path of $O_i$ may not end at $O_i$. We now formally have:
\begin{theorem1} \label{thm_EEL}
Under LiNGAM and d-separation faithfulness, if at most $d$ observed variables lie on a confounding path of any member of $\bm{O}$, then EEL with $l\leq d$ recovers the inducing terms $\bm{E}^*$.
\end{theorem1}
\noindent EEL thus partials out variables at each iteration and then discovers all inducing terms by only searching over subsets of variables adjacent in $\mathcal{G}$. In contrast, algorithms that discover causal structure in the confounded setting, such as the FCI, do not partial out variables but must search over an often much larger set of variables that lie on sequences of (undirected) inducing paths \citep{Spirtes00,Zhang08}.

\begin{algorithm}[t]
\hspace*{\algorithmicindent} \textbf{Input:} $\bm{O}$\\
 \hspace*{\algorithmicindent}  \textbf{Output:}  $\bm{E}^*,\mathcal{G}$
 \begin{algorithmic}[1]
  \State $\mathcal{G} \leftarrow$ complete undirected graph over $\bm{O}$ \label{alg_EEL:graph}
 \State \hlc[gray!15]{$l \leftarrow 0$}
 \Repeat
 \State \hlc[gray!15]{$l=l+1$}
  \State $\bm{Y} \leftarrow \emptyset$
 \ForAll{$O_j$ s.t. $|\textnormal{Adj}_\mathcal{G}(O_j) |$ \hlc[gray!15]{$\geq l$}}
 \Repeat
 \State \hlc[gray!15]{Choose a new $\bm{W} \subseteq \textnormal{Adj}_\mathcal{G}(O_j) $ s.t. $|\bm{W}|=l$}
 \If{$R_{O_j\bm{W}} \ci O_i,$ \hlc[gray!15]{$\forall O_i \in \bm{W}$} \label{alg_EEL:indep}}
 \State $\bm{Y} \leftarrow \bm{W}; \textbf{break}$ \label{alg_EEL:V}
  \EndIf
\Until{all $\bm{W} \subseteq \textnormal{Adj}_\mathcal{G}(O_j) $ \hlc[gray!15]{with $|\bm{W}| = l$} have been considered}
 \If{ $\bm{Y} \not = \emptyset$}
        \State Partial out $\bm{Y}$ from $O_j$ \label{alg_EEL:partial}
        \State Remove $\bm{Y}$ from $\textnormal{Adj}_\mathcal{G}(O_j) $ \label{alg_EEL:adj}
        \State \hlc[gray!15]{$l \leftarrow 0$}; \textbf{break} \label{alg_EEL:reset}
        \EndIf
 \EndFor
 \Until{all vertices satisfy $|\textnormal{Adj}_\mathcal{G}(O_j) |$ \hlc[gray!15]{$< l$}}
 \State $\bm{E}^* \leftarrow \bm{O}$
\end{algorithmic}

 \caption{Extract Errors with Latents (EEL)} \label{alg_EEL}
\end{algorithm}

We must of course perform the necessary regressions and independence tests with $n$ samples in practice. We assume that the independence test requires $O(n \textnormal{log}(n))$ time \citep{Even20,Even21} and consider the standard $O(n^2d+d^3)$ complexity of linear regression. The outer and innermost loops of EEL iterate over at most $\sum_{k=1}^{d} { q-1 \choose k}$ combinations with an independence oracle, and the second loop over at most $q$ variables. EEL therefore depends polynomially on the number of variables $q$ because $O(q \sum_{k=1}^{d} { q-1 \choose k}) = O(q^{d+1})$. We conclude that EEL theoretically takes $O(q^{d+1}n^2d)$ time in the oracle setting. However, the independence tests take longer than linear regression in practice due to the existence of highly optimized linear algebra libraries, so EEL runs in $O(q^{d+1} n \textnormal{log}(n))$ time for realistic sample sizes.

\section{Causal \& Predictive Contributions} \label{sec:contribution} 

We want to quantify the sample-specific total effect of $E_i$ on $D$, but EEL can only recover the inducing terms $\bm{E}^*$ when confounding exists. The variable $E_i^*$ is a linear combination of $E_i$ and some of the error terms and latent variables that are ancestors of $O_i$ per Equation \eqref{eq:C_total}. The sample-specific total effect of $E_i^*$ can therefore lie far from that of $E_i$. Even worse, the abstract quantity $E_i^*$ may not correspond to any real-world entity that we can manipulate in practice.

We instead seek a unified variable importance measure that (1) identifies the sample-specific total effects of the error terms \textit{when possible} and (2) otherwise corresponds to a measure of predictivity rather than causality. The output of EEL must also clearly indicate when (1) or (2) holds. 

We in particular consider the following Shapley value as a natural generalization of Equation \eqref{eq_shapley}, where we have replaced $\bm{E}$ with $\bm{E}^*$:
\begin{equation} \label{eq:Shap_inducing}
    S_i^* = \frac{1}{q}\sum_{\bm{W} \subseteq (\bm{E}^* \setminus E^*_i)} \frac{1}{\binom{q-1} {|\bm{W}|}} \gamma_{E_i^* \bm{W}}.
\end{equation}
We can gain a deeper understanding of $S_i^*$ using the undirected graph $\mathcal{G}$ provided by EEL. 

EEL instantiates the graph $\mathcal{G}$ over $\bm{O}$ in Line \ref{alg_EEL:graph}, but the graph summarizes the dependence relations between the inducing terms $\bm{E}^*$ when EEL terminates. We can construct the final form of $\mathcal{G}$ using the sets $\bm{C}_i$ with the following procedure:
\begin{enumerate}[leftmargin=*]
\item Instantiate an empty graph $\mathcal{G}$ over $\bm{E}^*$;
\item Draw an undirected edge between $E_i^*$ and $E_j^*$ if and only if $\bm{C}_i \cap \bm{C}_j \not = \emptyset$ for all pairs $\{O_i,O_j\}$.
\end{enumerate}
By construction:
\begin{proposition1}
Two inducing terms are adjacent in $\mathcal{G}$ if and only if they involve a common error term or latent variable. 
\end{proposition1}
\noindent The graph $\mathcal{G}$ therefore implies only small groups of dependent inducing terms. Let $\bm{B}_i^*$ denote the inducing terms with corresponding vertices adjacent to $E_i^*$ in $\mathcal{G}$. Consider:
\begin{equation} \nonumber
\psi_k = \frac{q}{\binom{|\bm{B}_i^*|-1}{k}{|\bm{B}_i^*|}}.
\end{equation}
Let $\delta$ denote the vector of coefficients obtained by logistically regressing $D$ on $\bm{E}^*$ so that $E_i^* \delta_{i}=(\bm{C}_i\theta_{\bm{C}_i i}) \delta_{i}$. Then:
\begin{theorem1} \label{thm:shap}
The following relation holds under a linear model: $\gamma_{E_i^* \bm{W}} = E_i^* \delta_{i} - \mathbb{E}(E_i^* | \bm{V})\delta_{i},$ where $\bm{W} \subseteq (\bm{E}^* \setminus E_i^*)$ and $ \bm{V} = (\bm{B}_i^* \setminus E_i^*) \cap \bm{W}$, so that:
\begin{equation} \label{eq:Shap_final}
 S_i^* = E_i^* \delta_{i} - \frac{\delta_{i}}{q}  \sum\limits_{\bm{V}\subseteq (\bm{B}_i^* \setminus E_i^*) } \psi_{|\bm{V}|} \mathbb{E}(E_i^* | \bm{V}).
\end{equation}
\end{theorem1}
\noindent In other words, $S_i^*=E_i^*\delta_i = E_i \theta_{iD}$ when $\mathcal{G}$ has no adjacencies because $E_i^* = E_i$ and $\delta_i = \theta_{iD}$. $S_i^*$ thus corresponds to $S_i$ when $O_i$ has no adjacencies in $\mathcal{G}$ and to a measure of predictivity when $O_i$ has adjacencies in $\mathcal{G}$. Furthermore, $\bm{S}^*$ is a unified measure still uniquely satisfying the linearity, efficiency, symmetry and null player properties.

The above result also implies that we can compute the Shapley value using subsets of $\bm{B}_i^* \setminus E_i^*$ rather than subsets of the much larger set $\bm{E}^* \setminus E_i^*$ in Equation \eqref{eq:Shap_inducing}. We estimate the expectations in Equation \eqref{eq:Shap_final} quickly even when $q$ is large, so long as $|\bm{B}_i^*|$ is small (e.g., $|\bm{B}_i^*| \leq 10$). 

If $|\bm{B}_i^*|$ is also large, then we estimate $S_i^*$ by Monte Carlo, where we sample the error terms with probabilities obeying the Shapley weights. We first sample $K$ with probability $\mathbb{P}(K) = 1/|\bm{B}_i^*|$. We then sample a set $\bm{V}$ by choosing a random subset of $\bm{B}_i^* \setminus E_i^*$ with size $K=k$; in other words, we sample $\bm{V}$ with probability $1/\binom{|\bm{B}_i^*|-1}{k}$ uniformly. We do not need to resort to Monte Carlo for the vast majority of cases because $\mathcal{G}$ is sparse in practice.

\section{Experiments} \label{sec:exp}

We compared EEL against the following algorithms representing the state of the art:
\begin{enumerate}[leftmargin=*]
\setcounter{enumi}{1}
\item Root Causal Inference (RCI): extracts error terms from the top-down by regressing on root vertices using a localized version of DirectLiNGAM and then computes Shapley values \citep{Strobl22a}.
\item Generalized Root Causal Inference (GRCI): extracts error terms from the bottom-up by regressing on parents of sink vertices and then computes Shapley values \citep{Strobl22b}.
\item Independent Component Analysis (ICA): performs ICA to extract the independent error terms and ranks variables according to a random forest permutation measure \citep{Lasko19}.
\item Root Causal Analysis of Outliers (RCAO): defines root causes according to an outlier score and computes Shapley values using the outlier scores and an estimated DAG \citep{Budhathoki22}.
\item Model Substitution (MS): re-samples the underlying DAG after substituting causal conditionals in an estimated DAG \citep{Budhathoki21}.
\end{enumerate}
\noindent See Appendix \ref{app:RW} for a comprehensive review of related work. We fixed the Type I error rate of EEL to 0.05 and estimated the Shapley values using MARS regression \citep{Friedman91}. We further standardized the data to prevent gaming of the marginal variances \citep{Reisach21}.

We do not have access to the ground truth Shapley values due to the unknown conditional expectations in Equation \eqref{eq:Shap_final}. We therefore approximated the conditional expectations to high accuracy by
training a committee of ten Linear Model Trees \citep{Quinlan92} -- a different model class than MARS -- on a sample of one hundred thousand ground truth inducing terms. We otherwise used the ground truth inducing terms, total causal effects and dependence graph $\mathcal{G}$ to compute Equation \eqref{eq:Shap_final} for each $O_i \in \bm{O}$.

\textbf{Reproducibility.} All code and data for reproducing experimental results are available at \\github.com/ericstrobl/EEL.

\subsection{Synthetic Data}

\subsubsection{Data Generation} We generated linear structural equation models using the following procedure. We first created a DAG with $p=15$ variables and an expected neighborhood size of two by creating a random adjacency matrix with independent realizations from a Bernoulli$(2/(p-1))$ distribution in the upper triangle portion of the matrix. We chose $D$ uniformly from the set of vertices without children and at least one parent. We selected 0, 10 or 20\% of the vertices as unobserved confounders provided each had at least two observed children not including $D$ and no parents. We then replaced the ones in the matrix by independent realizations of a uniform distribution on $[-1,-0.25] \cup [0.25, 1]$. We chose the distribution of each error term by uniformly sampling from the following set of possibilities:
the t-distribution with five degrees of freedom,
the chi-square distribution with three degrees of freedom, and the
uniform distribution on $-1$ to $1$. We finally drew instantiations of $D$ according to a Bernoulli random variable with probabilities obeying a logistic function per Assumption \ref{assump_terminal}. We repeated the above procedure 120 times for sample sizes of one, ten and one hundred thousand and latent variables of 0, 10 and 20\%. We therefore generated a total of $120 \times 3 \times 3 = 1080$ independent datasets.

\subsubsection{Evaluation Criteria}

The output of the five algorithms differ, but we can convert the output of each algorithm to a ranked list of variables. The top ranked variables should correspond to the true root causes with the largest Shapley values. We therefore first compared the algorithms using rank biased overlap (RBO), a well-established measure designed to compare two ranked lists of potentially differing lengths \citep{Webber10}:
\begin{equation} \nonumber
    \frac{1}{n} \sum_{k=1}^n \sum_{i=1}^{r_k} \widetilde{s}_i^k | \widehat{\mathcal{R}}_{1:i}^k \cap \mathcal{R}_{1:i}^k|/i,
\end{equation}
where $s_i^k$ denotes the true Shapley value of the variable $O_i$ for sample $k$, $\widetilde{s}_i^k = \frac{s_i^k}{\sum_{i=1}^{r_k} s_i^k}$ the Shapley values normalized to sum to one, and $r_k$ the total number of root causes for sample $k$. The notation $\mathcal{R}_{1:i}^k$ refers to the first $i$ variables in the ranking $\mathcal{R}^k$ for sample $k$. RBO increases monotonically with depth and weighs top ranks more heavily. The metric equals one when the top ranks coincide exactly with the true sample-specific root causes sorted in decreasing order by Shapley values, and zero when no overlap exists. Higher is therefore better.

We also compared the algorithms using the traditional mean squared error (MSE) to the true Shapley values:
\begin{equation} \nonumber
    \frac{1}{nq} \sum_{k=1}^n \sum_{i=1}^q (\widehat{s}_i^k - s_i^k)^2.
\end{equation}
where lower is better. If an algorithm only outputs Shapley values for a subset of variables, then we set the estimated Shapley values to zero for the excluded subset.

\subsubsection{Results}

\begin{table*}[t]
\centering
\captionsetup{justification=centering,margin=2cm}
\begin{tabular}{cc|cccccc} 
\hhline{========}
\rowcolor[HTML]{FFFFFF} 
\textit{l}                   & \textit{n} & EEL            & RCI            & GRCI           & ICA   & RCAO  & MS    \\ \hline
\rowcolor[HTML]{FFFFFF} 
                             & 1,000      & 0.850          & \textbf{0.918} & 0.883          & 0.713 & 0.652 & 0.662 \\
\rowcolor[HTML]{FFFFFF} 
0\%                          & 10,000     & 0.962          & \textbf{0.975} & \textbf{0.971} & 0.779 & 0.669 & 0.673 \\
\rowcolor[HTML]{FFFFFF} 
                             & 100,000    & \textbf{0.980} & \textbf{0.993} & \textbf{0.992} & 0.796 & 0.669 & 0.670 \\ \hline
\rowcolor[HTML]{FFFFFF} 
                             & 1,000      & 0.806          & \textbf{0.859} & 0.826          & 0.678 & 0.574 & 0.579 \\
\rowcolor[HTML]{EFEFEF} 
\cellcolor[HTML]{FFFFFF}10\% & 10,000     & \textbf{0.931} & 0.904          & 0.890          & 0.738 & 0.595 & 0.596 \\
\rowcolor[HTML]{EFEFEF} 
\cellcolor[HTML]{FFFFFF}     & 100,000    & \textbf{0.961} & 0.910          & 0.899          & 0.756 & 0.595 & 0.592 \\ \hline
\rowcolor[HTML]{FFFFFF} 
                             & 1,000      & \textbf{0.781} & \textbf{0.784} & 0.763          & 0.624 & 0.479 & 0.500 \\
\rowcolor[HTML]{EFEFEF} 
\cellcolor[HTML]{FFFFFF}20\% & 10,000     & \textbf{0.892} & 0.806          & 0.796          & 0.677 & 0.507 & 0.519 \\
\rowcolor[HTML]{EFEFEF} 
\cellcolor[HTML]{FFFFFF}     & 100,000    & \textbf{0.938} & 0.811          & 0.800          & 0.695 & 0.508 & 0.517\\
\hhline{========}
\end{tabular}
\caption{RBO results with the synthetic data. EEL achieved the highest mean RBO values with larger sample sizes as highlighted in gray.} \label{exp_synth:RBO}
\end{table*}

We summarize the RBO results for the synthetic data in Table \ref{results:synth}. Bolded values denote the best performance in each row according to one-sided paired t-tests at a Bonferroni corrected threshold of 0.05/6, since we compared a total of six algorithms. We present tables summarizing the MSE and timing results in Appendix \ref{app:add_res}. RBO and MSE results were similar. 

EEL achieved the best performance in terms of both RBO and MSE with confounding once sample sizes reached ten thousand. The margin continued to widen with increasing sample size and confounding degree. EEL outperformed the second best algorithm by a 15.6\% margin with $n=100,000$ and $l=20\%$. EEL therefore requires a sizable number of samples in order to achieve state of the art performance. 

EEL underperformed both RCI and GRCI without confounding. The margin however was small, and we cannot expect EEL to outperform algorithms explicitly designed for the unconfounded case. For example, RCI exploits certain local properties in unconfounded LiNGAM to significantly reduce the search space. We conclude that EEL remains competitive when no latent common causes exist. 

\subsection{Real Data}

\subsubsection{Diabetes}

We ran the algorithms on a real clinical dataset to identify \textit{patient}-specific root causes of diabetes. The dataset contains measurements of 8 variables related to the metabolic system among 768 patients of Pima Indian ancestry \citep{Smith88}.\footnote{https://www.kaggle.com/datasets/uciml/pima-indians-diabetes-database} Diabetes is a well-studied disease, so we asked a physician to generate the ground truth causal graph shown in the Appendix. We derived the values of the error terms via linear regression on the parents. We chose one to two latent variables uniformly at random from age and the diabetes pedigree function. The target is a binary diagnostic label of diabetes. 

\begin{figure}[t]
    \centering 
\begin{subfigure}{0.49\textwidth} \centering
\includegraphics[scale=0.65]{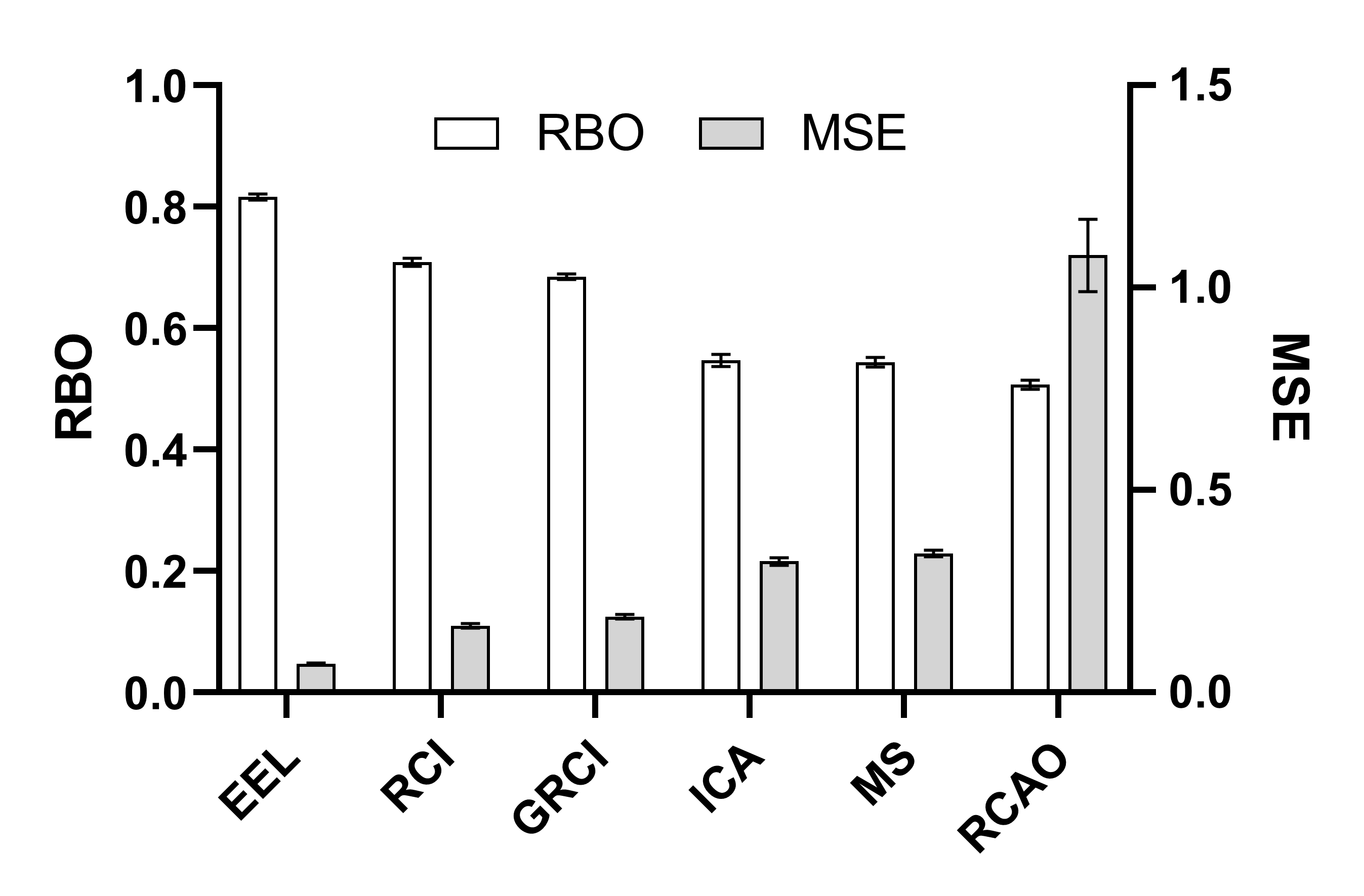}
\vspace{-5mm}
\caption{} \label{fig:Diabetes}
\end{subfigure}\begin{subfigure}{0.5\textwidth} \centering
\includegraphics[scale=0.65]{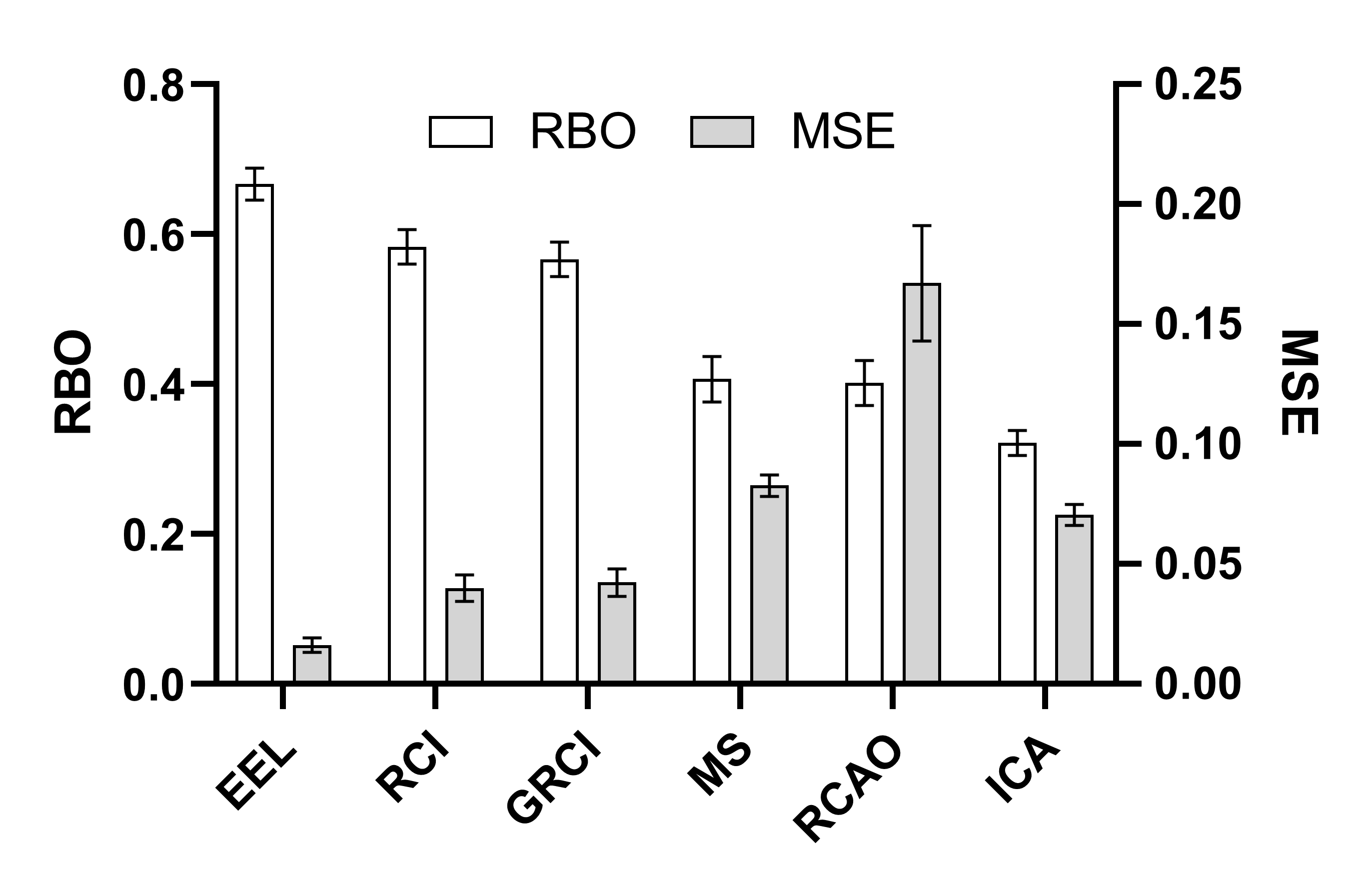}
\vspace{-5mm}
\caption{} \label{fig:Sachs}
\end{subfigure}
\caption{RBO and MSE results for the (a) diabetes and (b) flow cytometry datasets. Error bars denote 95\% confidence intervals. EEL again achieved the highest RBO and lowest MSE in both datasets.}  
\end{figure}

We summarize the results as averaged over 200 bootstrapped datasets in Figure \ref{fig:Diabetes} with algorithms sorted in decreasing order according to mean RBO value. EEL outperformed its nearest competitor by a 10.7 point RBO margin. EEL also achieved a 57.8\% reduction in the MSE. Both results were significant at a Bonferroni corrected threshold of 0.05/6 by paired t-tests. The algorithm completed in 9.2 seconds on average. We conclude that EEL achieves the best performance in this dataset.

\subsubsection{Flow Cytometry}
We next ran the algorithms on a real flow cytometry dataset to identify \textit{cell}-specific root causes. The dataset from  \citep{Sachs05} contains measurements of 11 phosphoproteins and phospholipids from 7466 primary human immune system cells across 9 experimental conditions.\footnote{https://arxiv.org/src/1805.03108v1/anc/data.txt} We log-transformed the data and standardized the samples in each experimental condition as recommended in \citep{Ramsey18}. We again derived the values of the error terms via linear regression on parents using the ground truth causal graph. We chose one to three latent variables uniformly at random from the options PKA, PKC and PIP3. We passed the mean of one to three observed variables, also chosen uniformly at random, through a logistic function for the binary target. We finally repeated the above process 200 times on bootstrapped samples.

We summarize the results in Figure \ref{fig:Sachs}. EEL outperformed all other algorithms by at least an 8.3 point margin according to RBO. EEL similarly achieved a 59.3\% reduction of the MSE from its nearest competitor. The algorithm took 64.6 seconds on average. We conclude that both real dataset results mimic those seen with the synthetic data.

\section{Conclusion}
We presented a novel algorithm called EEL that recovers the error terms of a structural equation model up to directed inducing paths. EEL also returns a sparse graph $\mathcal{G}$ summarizing the statistical dependencies between the recovered terms. We used the graph to quickly compute Shapley values, a unified measure corresponding to the sample-specific total effect when the inducing term $E_i^*$ corresponds to its associated error term $E_i$. Experiments demonstrated considerable improvements in accuracy relative to existing methods. We conclude that the combination of EEL and Shapley values offers a principled framework for performing sample-specific root causal inference with latent variables.

\acks{TBD}

\bibliography{biblio}

\section{Appendix}

\subsection{DirectLiNGAM} \label{app:DL}

The DirectLiNGAM (DL) algorithm is a well-known method for estimating the error terms assuming LiNGAM and no confounding where $\bm{X} = \bm{O}$ \citep{Shimizu11}. We will build upon DL in the next section, so we require a deep understanding of the algorithm's inner workings. More accurate and much faster variants of DL exist \citep{Strobl22a}, but we present the simplest version here to emphasize general concepts rather than algorithmic details.

DL capitalizes on the following result:
\begin{proposition1} \label{prop_noC}
    \citep{Shimizu11} Under LiNGAM and no confounding, $O_i$ is independent of the residuals $R_{O_jO_i}$ for \underline{all} $ O_j \in (\bm{O}\setminus O_i)$ if and only if $O_i=E_i$. Moreover, partialing out $O_i=E_i$ from $\bm{O}\setminus O_i$ generates another LiNGAM model.
\end{proposition1}
\noindent All error terms correspond to root vertices, or vertices without ancestors. The algorithm therefore extracts an error term in each iteration by performing a series of univariate regressions to identify a root vertex. Partialing out the root vertex then recovers another LiNGAM model with a new set of root vertices, so DL repeats the process until it recovers all error terms.

We summarize DL in more detail in Algorithm \ref{alg_DL}. The algorithm calls FindRoot in Line \ref{alg_DL:findRoot}, which we in turn summarize in Algorithm \ref{alg_FindRoot}. FindRoot regresses each variable $O_i \in \bm{O}$ onto each variable $O_j \in (\bm{O} \setminus O_i)$. The algorithm then determines whether the residuals $R_{O_jO_i}$ and $O_i$ are independent in Line \ref{alg_FindRoot:dir}, then vice versa in Line \ref{alg_FindRoot:rev_dir}, using the independence measure $\mathcal{I}_{ij}$; the non-negative measure equals zero if and only if independence holds \citep{Hyvarinen13}. FindRoot finally identifies the variable $O_i$ \textit{most} independent of its residuals in Line \ref{alg_FindRoot:id_root}, thus bypassing the need for formal hypothesis testing with a predetermined Type I error rate. DL then removes $O_i$ from consideration in Line \ref{alg_DL:remove} and replaces each $O_j$ by its residuals $R_{O_jO_i}$ in Line \ref{alg_DL:partial}. The algorithm iterates through this process until all variables in $\bm{O}$ have been replaced by their error terms. DL therefore ultimately outputs $\bm{E}$ as desired.

\begin{algorithm}[t]
 \hspace*{\algorithmicindent} \textbf{Input:} $\bm{O}$\\
 \hspace*{\algorithmicindent}  \textbf{Output:}  $\bm{E}$
\begin{algorithmic}[1]
\State $\bm{U} \leftarrow \bm{O}$
\Repeat
    \State  $G \leftarrow$ FindRoot($\bm{O},\bm{U}$) \label{alg_DL:findRoot}
    \State $\bm{U} \leftarrow \bm{U} \setminus G$ \label{alg_DL:remove}
    \State $(\bm{O} \setminus G) \leftarrow $ partial out $G$ from $\bm{O} \setminus G$ \label{alg_DL:partial}
\Until{$\bm{U} = \emptyset$}
\State $\bm{E} \leftarrow \bm{O}$
\end{algorithmic}
\caption{DirectLiNGAM (DL)} \label{alg_DL}
\end{algorithm}

\begin{algorithm}[t]
 \hspace*{\algorithmicindent} \textbf{Input:} $\bm{O},\bm{U}$\\
 \hspace*{\algorithmicindent}  \textbf{Output:}  root $G$
\begin{algorithmic}[1]
\State \textbf{return} $\bm{U}$  if $|\bm{U}| = 1$
\State $\bm{T} = \bm{0}_{|\bm{U}|}$
\For{$i \in [|\bm{U}|-1]$ \label{alg_FindRoot:iter_s}}
    \For{$j \in \{i+1, \dots, |\bm{U}|\}$}
        \State $T_i = T_i + \mathcal{I}_{ij}$ \label{alg_FindRoot:dir}
        \State $T_j = T_j + \mathcal{I}_{ji}$ \label{alg_FindRoot:rev_dir}
    \EndFor
\EndFor
\State $G \leftarrow \bm{U}[\argmin_i T_i]$ \label{alg_FindRoot:id_root}
\end{algorithmic}
\caption{FindRoot} \label{alg_FindRoot}
\end{algorithm}

\subsection{Integrating Hypothesis Testing} \label{app:EE}

DL unfortunately carries two main shortcomings:
\begin{enumerate}[leftmargin=*]
    \item The algorithm finds the variable most independent of its residuals in Line \ref{alg_FindRoot:id_root} of FindRoot. This process eliminates the need for hypothesis testing but ultimately slows down the algorithm by requiring that it check all pairs of variables in $\bm{U}$ in each iteration. 
    \item DL partials out the effect of each root vertex from \textit{all} remaining vertices because Equation \eqref{eq:SEM_lin} implies a linear relation from root vertices to their non-ancestors. We cannot apply this strategy to the confounded setting because $\bm{L}$ contains some of the root vertices.
\end{enumerate}
We rectify both of these issues with a new method called Extract Errors (EE) that also assumes no confounding. EE capitalizes on Lemma \ref{prop:C}, a simpler but analogous result to Proposition \ref{prop_noC}.

\begin{algorithm}[t]
\hspace*{\algorithmicindent} \textbf{Input:} $\bm{O}$\\
 \hspace*{\algorithmicindent}  \textbf{Output:}  $\bm{E}$
 \begin{algorithmic}[1]
 \State $\mathcal{G} \leftarrow$ complete undirected graph over $\bm{O}$ \label{alg_EE:graph}
 \ForAll{$O_j$ s.t. $|\textnormal{Adj}_\mathcal{G}(O_j) | >0$}
 \State $Y \leftarrow \emptyset$
 \Repeat
 \State Choose a new $O_i \in \textnormal{Adj}_\mathcal{G}(O_j) $
 \If{$ R_{O_jO_i} \ci O_i $} \label{alg_EE:ind}
 \State $Y \leftarrow O_i; \textbf{break}$ \label{alg_EE:V}
  \EndIf
 \Until{all vertices in $\textnormal{Adj}_\mathcal{G}(O_j)$ have been considered}
\If{ $Y \not = \emptyset$}
        \State Partial out $Y$ from $O_j$ \label{alg_EE:partial}
        \State Remove $Y$ from $\textnormal{Adj}_\mathcal{G}(O_j) $ \label{alg_EE:delete}
        \EndIf
\EndFor
\State $\bm{E} \leftarrow \bm{O}$
\end{algorithmic}
 \caption{ExtractErrors (EE)} \label{alg_EE}
\end{algorithm}

We can always find a variable $O_i \in \bm{O}$ where $O_i \ci R_{O_j\bm{W}}$ with $|\bm{W}| = 1$ in the unconfounded setting because we observe all root vertices (for instance, $\bm{W} = O_i$ when $O_i$ is a root vertex). We therefore set $|\bm{W}| = 1$ to focus on \textit{univariate} regressions. We consider a new algorithm called Extract Errors (EE) in Algorithm \ref{alg_EE}. The algorithm first instantiates a complete undirected graph $\mathcal{G}$ over $\bm{O}$ in Line \ref{alg_EE:graph}. EE then uses hypothesis testing to identify the variable $O_i \in \textnormal{Adj}_{\mathcal{G}}(O_j)$ independent of the residuals $R_{O_jO_i}$ in Line \ref{alg_EE:ind}. Subsequently, EE partials out $O_i$ from $O_j$ in Line \ref{alg_EE:partial} and eliminates the corresponding adjacency from $\mathcal{G}$ in Line \ref{alg_EE:delete}. The algorithm terminates once $\mathcal{G}$ contains no adjacencies -- i.e., once EE partials out all ancestral relationships. The correctness of EE follows as a corollary of Theorem \ref{thm_EEL}:
\begin{corollary1} \label{cor:EE}
Under LiNGAM and d-separation faithfulness, if no confounding exists, then EEL with $l \leq 1$ recovers the error terms $\bm{E}$.
\end{corollary1}

\begin{tcolorbox}[breakable,enhanced,frame hidden]
Line \ref{alg_EE:ind} represents the key step of EE because it allows the algorithm to select the first $O_i$ inducing residuals independent of \textit{some} $O_j \in (\bm{O} \setminus O_i)$, as opposed to \textit{all} $O_j \in (\bm{O} \setminus O_i)$ like in DL. EE can therefore quickly partial out a variable even if it is \textit{not} a root vertex in Line \ref{alg_EE:partial}. This property will come in handy when we introduce confounders because we may not observe a root vertex when confounders exist.
\end{tcolorbox}

\subsection{Related Work} \label{app:RW}

EEL is closely related to several lines of work. \cite{Lasko19} proposed a methodology of extracting the error terms of an SEM using ICA, although the authors did not connect the approach to causality. We previously proposed methods for identifying sample-specific root causes in the linear and non-linear settings \citep{Strobl22a,Strobl22b}. All of these methods however assume no confounding, whereas EEL accounts for unobserved variables by recovering inducing terms.

Our work is more broadly related to a suite of root causal analysis methodologies that identify sample-specific root causes in industrial or healthcare applications \citep{Anderson06,Wu08}. However, these methods take a painstaking manual approach that either implicitly or explicitly generates the underlying causal graph. Most strategies also focus on identifying human errors in \textit{man-made} systems with well-understood causal processes. We on the other hand focus on identifying biological errors in nature or, more generally, errors where the underlying causal process is largely unknown and difficult to understand.

Other computational approaches also exist for identifying root causes, but they again assume a known or estimated causal graph. For example, a recent paper introduced a method called Root Causal Analysis of Outliers (RCAO) that considers the root causes of outlier events \citep{Budhathoki22}. RCAO however assumes that the user can recover the error terms even in the confounded setting. The algorithm also redraws the values of the error terms and therefore identifies root causes at the population rather than at the sample-specific level. Finally, RCAO assumes that the label corresponds to an outlier event with a noiseless cut-off, even though the cut-off score for a diagnosis is noisy because it depends on the diagnostician in practice. The Model Substitution (MS) method proposed in \citep{Budhathoki21} makes similar assumptions. Another strategy allows a noisy cut-off score but requires paired rather than more widely available case control data \citep{Budhathoki22b}. EEL instead (1) utilizes case control data, (2) discovers the error terms directly without recovering or accessing the underlying causal graph, (3) identifies root causes at the sample-specific level and (4) allows a noisy diagnostic label with the logistic link. EEL is therefore more suitable for the biomedical setting with complex disease.

\subsection{Proofs} \label{app:proofs}

\begin{lemma1} (Darmois-Skitovitch) \label{lem:DS}
Suppose we can represent two random variables $O_1$ and $O_2$ as linear combinations of the mutually independent terms in $\bm{T}$:
\begin{equation} \nonumber
O_1 = \sum_{i=1}^p T_i \theta_{i1}  \textnormal{ and } O_2 = \sum_{i=1}^p T_i \theta_{i2}.
\end{equation}
If some $T_j$ for which $\theta_{j1} \theta_{j2} \not = 0$ is non-Gaussian, then $O_1$ and $O_2$ are dependent.
\end{lemma1}

Let $\bm{W}$ denote a set of arbitrary linear combinations of error terms and latent variables in a minimal set $\bm{S} \subseteq \bm{T}$ for the proposition below.
\begin{repproposition1}{prop:C}
    Under LiNGAM, $W_i$ is independent of the residuals $R_{O_j \bm{W}}$ for all $W_i \in \bm{W}$ if and only if $O_j$ can be written as a linear function of $\bm{W}$ \ul{plus} a linear function of $\bm{T} \setminus \bm{S}$. Thus, the residuals are a linear function of $\bm{T} \setminus \bm{S}$.
\end{repproposition1}
\begin{proof}
For the forward direction, if $W_i \ci R_{O_j \bm{W}}$ for all $W_i \in \bm{W}$, then $W_i$ and $R_{O_j \bm{W}}$ are linear combinations of non-overlapping subsets of $\bm{T}$ for all $W_i \in \bm{W}$ by Lemma \ref{lem:DS} under LiNGAM. This implies that $R_{O_j \bm{W}}$ is a linear function of $\bm{T} \setminus \bm{S}$, so $O_j$ must be a linear function of $\bm{W}$ plus a linear function of $\bm{T} \setminus \bm{S}$. For the backward direction, if $O_j$ can be written as a linear function of $\bm{W}$ plus a linear function of $\bm{T} \setminus \bm{S}$, then $R_{O_j \bm{W}}$ is a linear function of $\bm{T} \setminus \bm{S}$ only under LiNGAM. Hence $W_i \ci R_{O_j \bm{W}}$ for all $W_i \in \bm{W}$.
\end{proof}

\begin{lemma1} \citep{Strobl19} \label{lem:IP_or}
Under d-separation faithfulness, there exists an inducing path between $X_i$ and $X_j$ if and only if $X_i \not \ci X_j | \bm{W}$ for all $\bm{W} \subseteq \bm{O} \setminus \{X_i, X_j\}$.
\end{lemma1}

\begin{replemma1}{lem:IPs}
Under LiNGAM and d-separation faithfulness, if some entry in $\bm{W} \subseteq \bm{F} \setminus F_i$ corresponds to an observed vertex lying on a directed inducing path to $O_i$, then $R_{O_i \bm{W}} \not \ci F_j$ for some $F_j \in \bm{W}$.
\end{replemma1}
\begin{proof}
Let $F_k \in \bm{W}$ denote an entry lying on a directed inducing path to $O_i$. The directed inducing path $\Pi$ must contain at least one non-collider in $\bm{L}$, lest $\Pi$ induce a cycle in the DAG. Let $L_r$ denote one such non-collider that is also a latent parent of $O_k$. Let $\bm{A} \subseteq \bm{O}$ contain the observed ancestors of $O_i$ also corresponding to entries in $\bm{W}$. For example, $\bm{A}$ includes $O_k$ because $O_k$ is an ancestor of $O_i$, and $F_k \in \bm{W}$. Note that $E_k + L_r \gamma_{rk}$ is an additive component of $F_k$ and $E_k\theta_{E_ki} + L_r\theta_{L_ri} = E_k\theta_{E_ki} + L_r(\gamma_{r\bm{A}}\theta_{E_{\bm{A}}i} + \delta)$ is an additive component of $O_i$ by d-separation faithfulness. Assume $\delta = 0$ so that $\theta_{L_ri} = \gamma_{L_r\bm{A}}\theta_{E_{\bm{A}}i}$. But then $L_r \ci O_i | \bm{A}$ which contradicts the fact that $L_r \not \ci O_i | \bm{A}$ by the existence of an inducing path between $L_r$ and $O_i$ according to Lemma \ref{lem:IP_or} under d-separation faithfulness. We thus have $\delta \not = 0$. But then we cannot partial out all of the entries in $\bm{W}$ corresponding to $\bm{A}$ from $O_i$, so $R_{O_i \bm{W}} \not \ci F_j$ for some $F_j \in \bm{W}$.
\end{proof}

\begin{reptheorem1}{thm_EEL}
Under LiNGAM and d-separation faithfulness, if at most $d$ observed variables lie on a confounding path of any member of $\bm{O}$, then EEL with $l\leq d$ recovers the inducing terms $\bm{E}^*$.
\end{reptheorem1}
\begin{proof}
We prove the statement by induction. Base: suppose that only one vertex exists in $\bm{O}$. Then $E_i^* = E_i = O_i$, so EEL terminates with $E_i^* = O_i$. 

Step: suppose that EEL recovers $\bm{E}^*$, when there are $q$ variables in $\bm{O}$. We need to prove the statement when there are $q+1$ variables in $\bm{O}$. 
Without loss of generality, choose $O_{q+1}$ such that it is either an observed root vertex or a child of only an error term and latent variables so that $O_{q+1} = E^*_{q+1} = F_{q+1}$. We have two cases for any descendant $O_l$ of $O_{q+1}$:
\begin{itemize}[leftmargin=*]
\item $E_{q+1}$ lies on a directed inducing path to $O_l$. EEL cannot partial out $O_{q+1}$ from $O_l$ by Line \ref{alg_EEL:indep} and Lemma \ref{lem:IPs}. 
\item $E_{q+1}$ does not lie on a directed inducing path to $O_l$. Consider the largest set $\bm{U} \subseteq \bm{L}$ lying on a confounding path of $O_l$ from $E_{q+1}$, where every collider is an ancestor of $O_l$ and every non-collider is in $\bm{L}$ (the path may however not end at $O_l$). The at most $d$ children of $\bm{U}\cup E_{q+1}$ on the path are all ancestors of $O_l$. Now place these observed children in $\bm{Y}$. Then EEL partials out $\bm{Y}$ from $O_l$ in Line \ref{alg_EEL:partial} when $l\leq d$.
\end{itemize}
We chose $O_l$ as an arbitrary descendant of $O_{q+1}$, so we may repeat the above process for all descendants of $O_{q+1}$. 

Next, for each $O_m$ that is a child of $O_{q+1}$, set $E_m \leftarrow E_m + E_{q+1}\beta_{(q+1)m}$ and set $\gamma_{km} \leftarrow \gamma_{km} + \gamma_{k(q+1)}\beta_{(q+1)m}$ for each $L_k \in \textnormal{Pa}(O_{q+1}) \cap \bm{L}$. We finally eliminate $O_{q+1}$. The conclusion follows by the inductive hypothesis.
\end{proof}

\begin{repcorollary1}{cor:EE}
Under LiNGAM and d-separation faithfulness, if no confounding exists, then EEL with $l \leq 1$ recovers the error terms $\bm{E}$.
\end{repcorollary1}
\begin{proof}
At most one observed variable lies on a confounding path of $O_i$ -- that is, only $O_i$ itself -- so invoke Theorem \ref{thm_EEL} with $d = 1$. Observe further that $E_i$ is the only member of $\bm{T}$ that lies on a directed inducing path to any $O_i \in \bm{O}$. As a result, EEL with $l=1$ recovers $E_i^* = E_i$ for any $O_i \in \bm{O}$.
\end{proof}

\begin{reptheorem1}{thm:shap}
The following relation holds under a linear model: $\gamma_{E_i^* \bm{W}} = E_i^* \delta_{i} - \mathbb{E}(E_i^* | \bm{V})\delta_{i},$ where $\bm{W} \subseteq (\bm{E}^* \setminus E_i^*)$ and $ \bm{V} = (\bm{B}_i^* \setminus E_i^*) \cap \bm{W}$, so that:
\begin{equation} \nonumber
 S_i^* = E_i^* \delta_{i} - \frac{\delta_{i}}{q}  \sum\limits_{\bm{V}\subseteq (\bm{B}_i^* \setminus E_i^*) } \psi_{|\bm{V}|} \mathbb{E}(E_i^* | \bm{V}).
\end{equation}
\end{reptheorem1}
\begin{proof}
We may write the following sequence for $\gamma_{E_i^* \bm{W}}$, where $\overline{\bm{W}} = (\bm{E}^* \setminus E_i^*) \setminus \bm{W}$ and $ \bm{V} = (\bm{B}_i^* \setminus E_i^*) \cap \bm{W}$:
\begin{equation} \nonumber
\begin{aligned}
\mathbb{E}[f(\bm{E}^*)| \bm{W} ]
=\hspace{1mm}& \int f(\bm{W},\overline{\bm{w}}) p(\overline{\bm{w}}|\bm{W}) ~d\overline{\bm{w}}\\
=\hspace{1mm}& \int \Big(\sum_{e_i^* \in \overline{\bm{W}}} e_i^* \delta_{i} + \sum_{E_i^* \in \bm{W}} E_i^* \delta_{i} \Big) p(\overline{\bm{w}}|\bm{W}) ~d\overline{\bm{w}}\\
=\hspace{1mm}&\sum_{E_i^* \in \overline{\bm{W}}} \delta_{i} \int e_i^* p(e_i^*|\bm{W}) ~de_i^* + \sum_{E_i^* \in \bm{W}} E_i^* \delta_{i} \int p(\bm{W}) ~d\bm{W}\\
=\hspace{1mm}&\sum_{E_i^* \in \overline{\bm{W}}} \delta_{i} \mathbb{E}(E_i^*|\bm{V}) + \sum_{E_i^* \in \bm{W}} E_i^* \delta_{i}.
\end{aligned}
\end{equation}
We finally arrive at $\gamma_{E_i^* \bm{W}}$ by subtraction:
\begin{equation} \nonumber
\mathbb{E}[f(\bm{E}^*)| E_i^*,\bm{W}] - \mathbb{E}[f(\bm{E}^*)| \bm{W}] 
=\hspace{1mm} E_i^* \delta_{i} - \mathbb{E}(E_i^* | \bm{V})\delta_{i}.
\end{equation}

For $S_i^*$, we multiply (1) the number of sets $\bm{V} \subseteq (\bm{B}_i^* \setminus E_i^*)$ with $|\bm{V}|=k$ by (2) the Shapley weights to obtain:
\begin{equation} \nonumber
\underbrace{\binom{|\bm{B}_i^*|-1}{k} \sum_{j=0}^{q-|\bm{B}_i^*|} \binom{q-|\bm{B}_i^*|}{j}}_{(1)}\underbrace{\frac{1}{\binom{q-1}{j+k}}}_{(2)}
=\hspace{1mm}\frac{q}{|\bm{B}_i^*|}.
\end{equation}
We identify (1) by choosing $k$ elements from $\bm{B}_i^* \setminus E_i^*$ and then choosing the remaining elements from $(\bm{E}^* \setminus E_i^*) \cup (\bm{B}_i^* \setminus E_i^*)$. The equality follows by applying the creative telescoping algorithm \citep{Zeilberger91}. We then have:
\begin{equation} \nonumber
 S_i^* = E_i^* \delta_{i} - \frac{\delta_{i}}{q}  \sum_{k=0}^{|\bm{B}_i^*|-1}\sum\limits_{\substack{\bm{V}\subseteq (\bm{B}_i^* \setminus E_i^*) \\ |\bm{V}| = k}} \psi_k \mathbb{E}(E_i^* | \bm{V}),
\end{equation}
whence the conclusion follows.
\end{proof}

\subsection{Additional Results} \label{app:add_res}
\begin{table}[h]
\begin{subtable}{0.95\textwidth}  
\centering
\captionsetup{justification=centering,margin=2cm}
\begin{tabular}{cc|cccccc}
\hhline{========}
\rowcolor[HTML]{FFFFFF} 
\textit{l}                   & \textit{n} & EEL            & RCI                                    & GRCI                                   & ICA   & RCAO  & MS    \\ \hline
\rowcolor[HTML]{FFFFFF} 
                             & 1,000      & 0.027          & \textbf{0.008}                         & 0.013                                  & 0.199 & 0.566 & 0.206 \\
\rowcolor[HTML]{FFFFFF} 
0\%                          & 10,000     & 0.002          & \textbf{0.001}                         & \textbf{0.001}                         & 0.193 & 0.250 & 0.209 \\
\rowcolor[HTML]{FFFFFF} 
                             & 100,000    & \textbf{0.006} & \textbf{0.000}                         & \cellcolor[HTML]{FFFFFF}\textbf{0.000} & 0.189 & 0.191 & 0.209 \\ \hline
\rowcolor[HTML]{FFFFFF} 
                             & 1,000      & 0.075          & \textbf{0.050}                         & 0.063                                  & 0.268 & 0.686 & 0.280 \\
\rowcolor[HTML]{EFEFEF} 
\cellcolor[HTML]{FFFFFF}10\% & 10,000     & \textbf{0.010} & 0.032                                  & 0.033                                  & 0.260 & 0.323 & 0.284 \\
\rowcolor[HTML]{EFEFEF} 
\cellcolor[HTML]{FFFFFF}     & 100,000    & \textbf{0.005} & 0.032 & 0.035                                  & 0.256 & 0.266 & 0.286 \\ \hline
\rowcolor[HTML]{FFFFFF} 
                             & 1,000      & \textbf{0.147} & \textbf{0.159}                         & 0.180                                  & 0.407 & 1.015 & 0.422 \\
\rowcolor[HTML]{EFEFEF} 
\cellcolor[HTML]{FFFFFF}20\% & 10,000     & \textbf{0.047} & 0.137                                  & 0.147                                  & 0.399 & 0.491 & 0.427 \\
\rowcolor[HTML]{EFEFEF} 
\cellcolor[HTML]{FFFFFF}     & 100,000    & \textbf{0.023} & 0.140                                  & 0.142                                  & 0.394 & 0.408 & 0.432\\
\hhline{========}
\end{tabular}
\caption{MSE}
\end{subtable}

\vspace{5mm}
\begin{subtable}{0.95\textwidth}  
\centering
\captionsetup{justification=centering,margin=2cm}
\begin{tabular}{cc|cccccc}
\hhline{========}
\textit{l} & \textit{n} & EEL   & RCI   & GRCI   & ICA   & RCAO  & MS    \\ \hline
           & 1,000      & 2.686 & 0.031 & 0.228  & 0.872 & 2.396 & 0.611 \\
0\%        & 10,000     & 14.29 & 0.238 & 22.90  & 13.60 & 15.12 & 6.206 \\
           & 100,000    & 89.21 & 2.382 & 160.84 & 521.0 & 480.7 & 48.17 \\ \hline
           & 1,000      & 2.294 & 0.033 & 0.193  & 0.268 & 2.580 & 0.588 \\
10\%       & 10,000     & 14.45 & 0.280 & 16.34  & 0.260 & 13.56 & 5.506 \\
           & 100,000    & 96.68 & 3.143 & 93.77  & 213.7 & 424.4 & 41.52 \\ \hline
           & 1,000      & 1.798 & 0.035 & 0.131  & 0.407 & 2.264 & 0.521 \\
20\%       & 10,000     & 13.71 & 0.306 & 11.86  & 0.399 & 11.91 & 5.158 \\
           & 100,000    & 114.2 & 3.242 & 79.38  & 177.7 & 392.0 & 36.96\\
\hhline{========}
\end{tabular}
\caption{Time in seconds}
\end{subtable}
\caption{Results with the synthetic datasets in terms of mean (a) MSE and (b) time in seconds. EEL achieved the lowest MSE mean values with enough samples as highlighted in gray. However, EEL took more time to complete than RCI.} \label{results:synth}
\end{table}

We refer to Table \ref{results:synth}. EEL never came in first or last in terms of timing. The mean time increased most notably with sample size but remained within $O(n \textnormal{log}(n))$ as expected per the complexity analysis in Section \ref{sec:EEL}. EEL only suffered a modest increase in time with higher degrees of confounding. We conclude that sample size drove most of the runtime of EEL in our experiments.

\subsection{Diabetes Graph} \label{app:diabetes}

\begin{figure}[h]
\centering
\begin{tikzpicture}[scale=1.0, shorten >=1pt,auto,node distance=2.8cm, semithick,
  inj/.pic = {\draw (0,0) -- ++ (0,2mm) 
                node[minimum size=2mm, fill=red!60,above] {}
                node[ minimum width=2mm, minimum height=5mm,above] (aux) {};
              \draw[thick] (aux.west) -- (aux.east); 
              \draw[thick,{Bar[width=2mm]}-{Hooks[width=4mm]}] (aux.center) -- ++ (0,4mm) coordinate (-inj);
              }]
                    
\tikzset{vertex/.style = {inner sep=0.4pt}}
\tikzset{edge/.style = {->,> = latex'}}
 
\node[vertex] (1) at  (0,0) {age};
\node[vertex] (2) at  (2,0) {pedigree};
\node[vertex] (3) at  (-1,-1) {preg};
\node[vertex] (4) at  (1,-1) {BMI};
\node[vertex] (5) at  (0,-2) {BP};
\node[vertex] (6) at  (3,-1) {glucose};
\node[vertex] (7) at  (3,-2) {insulin};
\node[vertex] (8) at  (1.5,-2) {skin};

\draw[edge] (1) to (3);
\draw[edge] (1) to (4);
\draw[edge] (1) to (5);
\draw[edge] (4) to (5);
\draw[edge] (2) to (4);
\draw[edge] (2) to (6);
\draw[edge] (6) to (7);
\draw[edge] (4) to (8);
\draw[edge] (1) to (6);
\draw[edge,bend right=40] (2) to (7);
\end{tikzpicture}

\end{figure}

\end{document}